\documentclass[letterpaper, 10 pt, journal, twoside]{ieeetran}

\IEEEoverridecommandlockouts                              
\usepackage{cite}
\usepackage{amsmath,amssymb,amsfonts}
\usepackage{algorithmic}
\usepackage{graphicx}
\usepackage{textcomp}
\usepackage{gensymb}
\usepackage{xcolor}
\usepackage{color}
\usepackage{wasysym}
\usepackage{multirow}		
\usepackage{array}										
\usepackage{colortbl}
\usepackage[per-mode=symbol]{siunitx} 
\usepackage[english]{babel}	
\usepackage[hidelinks]{hyperref}    %
\usepackage{blindtext}
\usepackage[ruled, linesnumbered]{algorithm2e}
\usepackage{pifont}

\usepackage{booktabs}
\usepackage{footmisc}

\usepackage{dblfloatfix}
\usepackage[T1]{fontenc}
\usepackage[utf8]{inputenc}
\title{Learning-based Nonlinear Model Predictive Control of Articulated Soft Robots using\\Recurrent Neural Networks
}

\author{Hendrik Schäfke$^{*}$, Tim-Lukas Habich$^{*}$, Christian Muhmann, \\Simon F. G. Ehlers,  Thomas Seel, and Moritz Schappler
\thanks{Manuscript received July 18, 2024; revised September 30, 2024; accepted October 26, 2024.}
\thanks{This paper was recommended for publication by editor Y.-L. Park upon evaluation of the associate editor and reviewers' comments.
	This work was supported by the Deutsche Forschungsgemeinschaft (DFG, German Research Foundation) -- 433586601 and INST 187/742-1 FUGG.} 
\thanks{All authors are with the Leibniz University Hannover, Institute of Mechatronic Systems, 30823 Garbsen, Germany,
	{\tt\footnotesize{\href{mailto:tim-lukas.habich@imes.uni-hannover.de}{tim-lukas.habich@imes.uni-hannover.de}}}}%
\thanks{$^{*}$Both authors contributed equally to this publication.}%
\thanks{Digital Object Identifier (DOI): see top of this page.}
}

\newif\ifcopyright
	\copyrighttrue

\newif\ifhighlightchanges

\ifhighlightchanges
\newcommand{\highlightredsec}[1]{\textcolor{red}{#1}}
\else
\newcommand{\highlightredsec}[1]{#1}
\fi
\newcommand{\highlightred}[1]{#1}
\newcommand\CPP{C\nolinebreak[4]\hspace{-.05em}\raisebox{.4ex}{\relsize{-3}{\textbf{++}}}}

\newcommand{\mm}[1]{\boldsymbol{#1}}

\newcommand{\mc}[1]{\mathcal{#1}}
\newcommand{\ind}[1]{\mathrm{#1}}
\newcommand{\R}{\mathbb{R}}
\usepackage[mathscr]{euscript}

\newcommand{\transpose}{^\mathrm{T}}
\definecolor{Gray}{gray}{0.85}

\newcolumntype{M}[1]{>{\centering\arraybackslash}m{#1}}
\newcolumntype{N}{@{}m{0pt}@{}}
\makeatletter
\newcommand{\removelatexerror}{\let\@latex@error\@gobble}
\makeatother

\graphicspath{{./images/}}

\begin{document}
\ifcopyright
{\LARGE IEEE Copyright Notice}
\newline
\fboxrule=0.4pt \fboxsep=3pt

\fbox{\begin{minipage}{1.1\linewidth}  
		\textcopyright\,\,2024\,\,IEEE. Personal use of this material is permitted. Permission from IEEE must be obtained for all other uses, in any current or future media, including reprinting/republishing this material for advertising or promotional purposes, creating new collective works, for resale or redistribution to servers or lists, or reuse of any copyrighted component of this work in other works. \\
		
		Accepted to be published in: IEEE Robotics and Automation Letters (RA-L), 2024.\\
		
		DOI: 10.1109/LRA.2024.3495579
\end{minipage}}
\else
\fi
\maketitle

\markboth{IEEE Robotics and Automation Letters. Preprint Version. Accepted October, 2024}
{Schäfke \MakeLowercase{\textit{et al.}}: Learning-based NMPC of Articulated Soft Robots using Recurrent Neural Networks} 

\begin{abstract}
Soft robots pose difficulties in terms of control, requiring novel strategies to effectively manipulate their compliant structures.
Model-based approaches face challenges due to the high dimensionality and nonlinearities such as hysteresis effects.
In contrast, learning-based approaches provide nonlinear models of different soft robots based only on measured data.
In this paper, recurrent neural networks~(RNNs) predict the behavior of an articulated soft robot~(ASR) with five degrees of freedom~(DoF).
RNNs based on gated recurrent units~(GRUs) are compared to the more commonly used long short-term memory~(LSTM) networks and show better accuracy.
The recurrence enables the capture of hysteresis effects that are inherent in soft robots due to viscoelasticity or friction but cannot be captured by simple feedforward networks.
The data-driven model is used within a nonlinear model predictive control~(NMPC), whereby the correct handling of the RNN's hidden states is focused.
A training approach is presented that allows measured values to be utilized in each control cycle.
This enables accurate predictions of short horizons based on sensor data, which is crucial for closed-loop NMPC.
The proposed learning-based NMPC enables trajectory tracking with an average error of \SI{1.2}{\degree} in experiments with the \highlightred{pneumatic five-DoF ASR}.
\end{abstract}
\begin{IEEEkeywords}
	Modeling, control, and learning for soft robots, machine learning for robot control, optimization and optimal control
\end{IEEEkeywords}
%
\section{Introduction}\label{sec:introduction}
\IEEEPARstart{I}{nspired} by nature, soft robots are revolutionizing the field of robotics due to their diverse designs and inherent compliance.
\highlightred{This enables safe human-robot interaction and integration into environments otherwise unsuitable for conventional rigid robots as softness causes less harm to their environment~\cite{Rus.2015}.}
For instance, an inflatable humanoid robot can interact intrinsically safe with humans compared to traditional rigid robots~\cite{Best.2015}.
In addition, compact soft robots allow them to be maneuvered into hard-to-reach areas.
\highlightred{However, challenges arise during modeling with conventional approaches due to complex geometries and nonlinearities, such as friction or air compressibility~\cite{Xavier.2022}.}
The viscoelastic (\mbox{time-,} temperature- and velocity-dependent) material results in strong hysteresis.
Combined with the many DoF, controlling such robots is more complex than other robot types~\cite{Santina.2023}.

\begin{figure}[] 
	\centering
	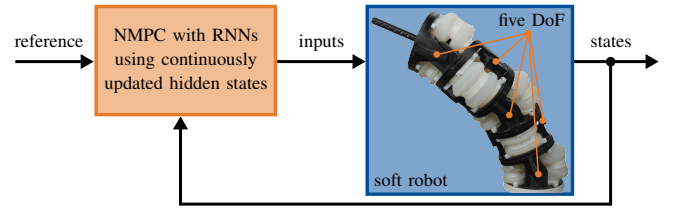
	\caption{Learning-based NMPC of a five-DoF ASR. The dynamic behavior is learned with recurrent neural networks and used as a dynamic constraint.} 
	\label{fig:cover}
	\vspace{-2mm}
\end{figure}

Interest in \textit{learning-based} modeling approaches for soft-robot control has increased, as they have shown promise in mitigating the given challenges.
\highlightred{By relying only on input/output data, model-based controllers can be implemented quickly through black-box learning.}
\highlightred{In the event of system changes (e.g., wear/replacement of soft materials or the addition of new actuators), only new data is needed to retrain the networks.}
Moreover, this approach is not dependent on a specific robot and can be transferred to different systems~\cite{Laschi.2023}.

We propose a learning-based model predictive control~(MPC) \highlightred{for} a multi-DoF soft robot as shown in Fig.~\ref{fig:cover}.
The robot dynamics are learned by RNNs based on LSTM cells and GRUs, which are used as a model in the nonlinear MPC.
This approach enables fast and simple soft-robot modeling and opens up the advantages of model-based optimal control~\cite{Best.2016}.
Learning-based NMPC for controlling soft robots has only been applied by a few researchers~\cite{Laschi.2023} and in particular, the use of RNNs, which are able to capture hysteresis effects, has not been adequately explored at present. 
The remainder of this paper is as follows: 
After an overview of related work and our contributions, preliminaries are presented in Section~\ref{sec:preliminaries}.
Section~\ref{sec:lb_control} describes the modeling using LSTM cells and GRUs and the design of the learning-based NMPC.
This is followed by experiments with the real ASR in Section~\ref{sec:experiments} and conclusions in~Section~\ref{sec:conclusions}.

\subsection{Related Work}\label{sub:related work}
MPC is a promising approach for controlling soft robots~\cite{Thuruthel.2018}.
\highlightred{For position control of a pneumatic soft robot, both linear quadratic control and MPC were implemented~\cite{Best.2016}.} 
\highlightred{The model was developed using \textit{first principles} and simplified using rigid-body assumptions.}
With MPC, position and stiffness were controlled simultaneously by including pressure conditions in the model~\cite{Gillespie.2016}.
\highlightred{Modeling soft-robot dynamics requires significant effort and expertise, and is often inaccurate due to simplifications regarding nonlinearities.}
In contrast, \textit{learning-based approaches} approximate system properties via data.
These methods are relevant in soft robotics as they capture complex nonlinear behaviors more effectively than conventional methods~\cite{Kim.2021}.
Further, black-box learning with data is significantly faster to develop than conventional gray-box modeling using first principles~\cite{Gillespie.2018}.

\textit{Feedforward neural networks}~(FNNs) are suitable as universal function approximators and were first applied to feedforward control of soft extensible continuum manipulators in~\cite{Braganza.2007}.
Furthermore, data-driven MPC was developed for the position control of a single-DoF soft actuator using simple first-order Markov FNNs as the dynamic model~\cite{Gillespie.2018}.
Using a similar approach, the position control was extended to a more complex system with multiple DoF based on a nonlinear evolutionary MPC algorithm~\cite{Hyatt.2019}.
This algorithm was published in~\cite{Cheney.2024} together with a framework for learning dynamics as an open-source library.
\highlightred{Another architecture was developed in~\cite{Johnson.2021} by training a surrogate FNN on simulation data from a first-principles model, with a second FNN representing the error.
By combining both networks, an NMPC algorithm was developed for position control of a multi-DoF soft robot, requiring less data than purely data-based approaches.}
In \cite{Null.2024}, a data-driven MPC was set up for a multi-DoF hydraulically actuated robot with an FNN model.
In contrast to the previous approaches, the hyperparameters of the MPC were tuned automatically.
However, \textit{all} approaches presented so far have the limitation that the implemented FNNs cannot capture hysteresis, which occurs strongly in soft robots due to the viscoelastic material behavior and friction.

One strategy to deal with hysteresis effects is to approximate the system dynamics using \textit{recurrent neural networks}, which is explicitly declared as future-work direction in~\cite{Gillespie.2018,Hyatt.2019}.
These are ideal for learning time sequences~\cite{Lipton.2015} and suitable as an alternative to state-space models~\cite{Schuessler.2019}.
They consider \textit{past states through recurrent layers}, which is required to capture hysteresis and are therefore preferred for modeling soft-robot dynamics~\cite{Armanini.2023, Chen.2024}.
\highlightred{The dynamic model of a dielectric elastomer actuator based on LSTM units was proposed in~\cite{Xiao.2020}.}
LSTM models were also used in~\cite{Li.2022} to generate actuation inputs for soft-robot control.
Compared to FNNs, they showed better prediction accuracy for small networks.
\highlightred{A bidirectional LSTM controller for modular soft robots with varying module numbers was introduced in~\cite{Chen.2024rebuttal}.}
\highlightred{Different network architectures (FNNs and RNNs) are compared for system identification in~\cite{Ogunmolu.2016}.}
\highlightred{Therein, the models based on GRUs outperformed LSTM models on a soft-robot example.}
However, neither the hyperparameters of the networks were optimized nor MPC was realized.
For a robotic catheter, an LSTM-based motion controller was used in~\cite{Wu.2022} to capture hysteresis.
In~\cite{Thuruthel.2017}, the forward dynamics of a soft manipulator were learned using a nonlinear autoregressive exogenous model~(NARX).
Since computing time was too high for MPC, only an open-loop predictive control policy was implemented.
However, this open-loop control scheme is problematic due to model errors or external disturbances. 
To solve this problem, a closed-loop control was implemented in~\cite{Thuruthel.2019} using reinforcement learning.
For this purpose, the RNN was used to simulate the robot, and the closed-loop control policies were learned using a second FNN.
This was further developed in~\cite{Centurelli.2022} to realize closed-loop control even with a previously unknown payload.
With these approaches, however, the \textit{advantages of MPC} and thus optimal control, such as real-time optimization and constraint satisfaction, cannot be utilized.
In a different approach, a convolutional neural network is used to learn the hysteresis model of a pneumatic muscle~\cite{Zhang.2019}. 
However, no control was implemented.
We used Gaussian processes in previous work~\cite{Habich.2023} to realize a learning-based position and stiffness feedforward control of a soft actuator.
Neither hysteresis effects were modeled, nor was MPC used.

Data-driven MPC for simultaneous position and stiffness control was successfully realized for a single actuator via LSTM units~\cite{Luong.2021}.
Using automatic differentiation, a linearized state-space model of the network was formulated and then used in the linear MPC.
Although initializing the hidden states of LSTM or GRU networks is crucial for short prediction horizons, \textit{none of the previously mentioned works} has addressed their handling in detail.
However, this is especially relevant for MPC, where poorly initialized hidden states lead to a low modeling accuracy within short prediction horizons.

\subsection{Contributions}
According to Laschi et al.~\cite{Laschi.2023}, \textit{incorporating traditional control architectures into learning modules} is essential for further advancements in the field of soft robotics.
In line with this recent perspective, we combine MPC as a traditional control architecture with learned RNNs as dynamics models, demonstrating a synergistic effect for improved system control.
\highlightred{There is a lack of experimental results on data-driven MPC with RNNs~\cite{Jung.2023} and it is listed as an open challenge~\cite{Hyatt.2019, Johnson.2021, Salzmann.2023}.}
To date, there is only one work~\cite{Luong.2021} in the soft-robotics field that uses learning-based MPC with RNNs.
There, however, only a \textit{linearized model} based on LSTM units is used \highlightred{for \textit{linear MPC}}, and it is applied to a \textit{simple one-DoF actuator}.
To the best of our knowledge, no work considers NMPC of multi-DoF soft robots with RNNs. 
\highlightred{The hyperparameter optimization (HPO), which is crucial for RNN performance, has also not been sufficiently considered in this context.}
We fill the gaps with the following contributions: \textbf{1)}~\highlightred{comparison of RNNs based on GRUs and LSTM units including systematic HPO to model a multi-DoF soft robot}, \textbf{2)}~solving the initialization problem of the hidden states for small prediction horizons to realize learning-based NMPC using RNNs, \textbf{3)}~validation of the RNN-based NMPC with experiments using the real soft robot, and \highlightred{\textbf{4)}~open-source publication\footnote{\label{foot:sponge}\highlightred{\url{https://tlhabich.github.io/sponge/rnn_mpc}}} of the codebase for neural-network training, HPO and MPC of the robot. }
%
\section{Preliminaries}\label{sec:preliminaries}
\subsection{Recurrent Neural Networks}\label{sub:gru}
RNNs have an additional feedback loop compared to FNNs, which allows them to use information from previous inputs to influence the current output.
RNNs are, therefore, able to predict time-varying sequential data by incorporating causal relationships from the past.
\highlightred{The LSTM unit uses internal gates to regulate what information to retain or discard~\cite{Hochreiter.1997}.}
\highlightred{GRUs are an RNN type based on the LSTM cell.}
In comparison, they are simpler in design, as they only use update and reset gates instead of input, forget, and output gates~\cite{Cho.2014}.
\highlightred{Thus, GRUs enable faster training, prediction, and optimization in the NMPC.}
\highlightred{Both solve the vanishing and exploding gradient issues through their gate structure.}

\highlightred{The hidden states~$\mm{h}_{k}$ at the discrete time step $k$ are crucial for the prediction quality, as they capture temporal dependencies in sequential data}\footnote{LSTMs have both hidden and cell states, while GRUs only have hidden states. For simplicity, only hidden states are mentioned, referring to both.\label{hidden_state_footnote}}.
They are recursively passed on as inputs to the next time step, which indicates that RNNs don't also have to receive the current system's states~$\mm{x}_{k}$ as inputs.
The latter would be ignored by \textit{conventional trained RNNs}, since they rely on the given hidden states.
For MPC applications, however, we want to explicitly \textit{use measurement data in order to achieve feedback control}. 
For this, the RNN training must be adjusted, which is presented in~\ref{sub:modeling}.
\subsection{Soft-Robot Platform}\label{sub:soft_robot_platform}
\begin{figure}[bt]
	\vspace{2.5mm}
	\centering
	\resizebox{.82\linewidth}{!}{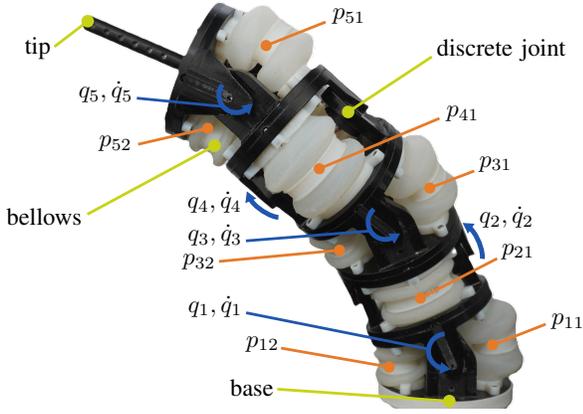}
	\caption{Soft-robot platform with $n=5$ discrete joints.} \label{fig:kinematic_chain_neu}
	\vspace{-2mm}
\end{figure}

The ASR used in this work is based on the semi-modular open-source design presented in~\cite{Habich.2024} and is shown in Figs.~\ref{fig:cover} and \ref{fig:kinematic_chain_neu}.
It consists of $n$ pneumatic soft actuators, which are stacked and alternately rotated by $\SI{90}{\degree}$ along the longitudinal axis.
Each discrete joint $i$ is actuated by air pressures~$\mm{p}_i{=}[p_{i1},p_{i2}]\transpose$ of antagonistically arranged bellows.
Each actuator contains a built-in Hall encoder for measuring the joint angles~$\mm{q}{=}[q_1,\dots,q_n]\transpose$.
The measured values are low-pass filtered and then numerically differentiated to obtain the joint velocities  $\dot{\mm{q}}$.
\highlightred{The desired bellows pressures~$\mm{p}_\ind{des}{=}[\mm{p}\transpose_\ind{1,des},\dots,\mm{p}\transpose_{n,\ind{des}}]\transpose$ are controlled using external proportional valves, which also measure the bellows pressures~${\mm{p}}{=}[\mm{p}\transpose_1,\dots,\mm{p}\transpose_n]\transpose$.}
Each joint angle $q_i$ results from the pressure difference between the antagonistic bellows.
Further information regarding the open-source platform can be found \highlightred{in the supplementary video of~\cite{Habich.2024}.
Only minor design improvements were made after publication,} namely a reduction in joint friction via smaller shaft diameters, less plastic deformation of the frames, thicker bellows for higher pressures, and larger tube diameters for faster pressure dynamics.
This improved version is also part of the corresponding website of~\cite{Habich.2024}.
%
\section{Learning-based Control of ASRs}\label{sec:lb_control}
This section presents the RNN training~(\ref{sub:modeling}) and the optimization of hyperparameters~(\ref{sub:hyperparam}).
Based on this, the learning-based NMPC is presented~(\ref{sub:lb_mpc}).

\subsection{Learning Robot Dynamics}\label{sub:modeling}
The ASR dynamics are approximated with GRUs and LSTM units.
Since they are very similar in structure, the following examples focus exclusively on the explanation of the GRU network. 
An ASR with $n{=}5$ joints is used for this paper, but this can easily be extended to additional actuators.
We denote $\boldsymbol{x}_{k}{=}[\mm{q}\transpose,\dot{\mm{q}}\transpose]\transpose$ as states and $\boldsymbol{u}_{k}{=}\mm{p}_{\ind{des}}$ as inputs of the dynamical system\footnote{The index $k$ is only used for $\mm{x}_k$, $\mm{u}_k$ and is omitted for $\mm{q},\dot{\mm{q}},\mm{p},\mm{p}_\ind{des}$.}. 
At the beginning of this research, it was considered to additionally use the measured pressures as states in order to better consider the pressure dynamics.
\highlightred{However, the valves' pressure control is fast enough so that the desired and measured pressures match closely with a time delay of $\SI{10}{\milli\second}{-}\SI{80}{\milli\second}$.}
Therefore, this did not significantly improve the prediction accuracy and, at the same time, increased the network's complexity, which would decrease the maximum possible NMPC frequency. 
Note that the pressure dynamics are still implicitly considered by using the above inputs and states during model learning.

The GRU training poses a time-series problem, whereby the prediction of future states $\hat{\boldsymbol{x}}_{k+1}$ can be expressed as
\begin{equation}\label{eq:neuralnetwork}
	[{\hat{\mm{x}}}_{k+1},\mm{h}_{k}]=\boldsymbol{f}(\boldsymbol{x}_{k},\boldsymbol{u}_{k}, \boldsymbol{h}_{k-1}),
\end{equation}
where $\boldsymbol{f}$ represents the RNN.
The GRUs must be able to accurately predict several time steps into the future \textit{using the current measured values in order to realize closed-loop NMPC}.
This behavior is imitated during the training of the neural networks.
A detailed description of the training procedure is given in Algorithm~\ref{alg:GRU_training} as pseudo-code.

\begin{figure}[]
	\removelatexerror
	\vspace{2mm}
	\begin{algorithm}[H]
		\caption{Training of RNNs compatible for MPC}\label{alg:GRU_training}
		{\small 
			\SetKwInOut{Input}{Input}
			\SetKwInOut{Output}{Output}
			\Input{$n_\ind{e},n_\ind{b},n_\ind{w},n_\ind{p},n_{\eta},\eta_{\ind{init}},\mm{X},\mm{Y}$}
			$\eta \gets \eta_\ind{init}$\;
			$\tilde{\mm{X}},\tilde{\mm{Y}} \gets$ Divide $\mm{X},\mm{Y}$ in sequences of length~$n_\ind{w}+n_\ind{p}$\;
			$\tilde{\mm{Y}} \gets$ delete first $n_\ind{w}$ points of each sequence\;
			$\mm{D}_\ind{t}, \mm{D}_\ind{v} \gets$ Split $\tilde{\mm{X}},\tilde{\mm{Y}}$ into training and validation data\;
			\ForEach{$\mathrm{epoch}\in\{1,\dots,n_\ind{e}\}$}{
				$\mc{L}_\ind{v,epoch} \gets0$\;
				$\mm{B}_\ind{t},\mm{B}_\ind{v} \gets$ Shuffle $\mm{D}_\ind{t},\mm{D}_\ind{v}$ and collect batches with~$n_\ind{b}$ sequences\;
						\ForEach{$\mm{B}\in\{\mm{B}_\ind{t},\mm{B}_\ind{v}\}$}{
							\ForEach{$\mm{b}\in\mm{B}$}{
								\highlightred{$\mm{b}_{\hat{Y}}\gets$ initialize empty list\;}
								\ForEach{$\mm{X}_\ind{seq}\in\mm{b}$}{
									$\mm{h}_{0}\gets$ initialize with zeros\;\label{warmupstart}				
									\ForEach{$k\in\{1,\dots,n_\ind{w}\}$}{
										$[{\hat{\mm{x}}}_{k+1},\mm{h}_{k}]=\boldsymbol{f}({\mm{x}}_{k},\mm{u}_{k},\mm{h}_{k-1})$\;
									}\label{warmupend}
									$\hat{\mm{x}}_{n_\ind{w}+1}\gets$initialize with $\mm{x}_{n_\ind{w}+1}$\;\label{predstart}
									\ForEach{$k\in\{n_\ind{w}{+}1,\dots,n_\ind{w}{+}1{+}n_\ind{p}\}$}{
										$[{\hat{\mm{x}}}_{k+1},\mm{h}_{k}]=\boldsymbol{f}(\hat{\mm{x}}_{k},\mm{u}_{k},\mm{h}_{k-1})$\;
										$\hat{\mm{Y}}_\ind{seq}\gets \hat{\mm{x}}_{k+1}$ add to sequence\;
									}\label{predend}
									$\mm{b}_{\hat{Y}}\gets\hat{\mm{Y}}_\ind{seq}$ add to batch;
								}
								
								\uIf{$\mm{B}=\mm{B}_\ind{t}$}{
									$\mc{L}_\ind{t} \gets \mathrm{MSE}(\mm{b}_{\hat{Y}}, \mm{b}_\ind{true})$\;
									$\mm{w}\gets$ Optimization with $\mc{L}_\ind{t},\eta$\;
								}
								\Else{$\mc{L}_\ind{v} \gets \mathrm{MSE}(\mm{b}_{\hat{Y}}, \mm{b}_\ind{true})$\;
									$\mc{L}_\ind{v,epoch} \gets$ sum up $\mc{L}_\ind{v}$\;}
						}}
						$\eta\gets\mathrm{ReduceLROnPlateau}(\mc{L}_\ind{v,epoch},n_{\eta})$\;\label{LRsched}
					}
				}
	\end{algorithm}
	\vspace{-1.5mm}
\end{figure}

\highlightred{The training requires the following inputs: number of epochs~$n_\ind{e}$, batch size~$n_\ind{b}$, warm-up steps~$n_\ind{w}$, prediction steps~$n_\ind{p}$, patience period for reducing the learning rate~$n_{\eta}$, initial learning rate~$\eta_{\ind{init}}$ and input data~$\mm{X}$ and $\mm{Y}$.
In general, $\mm{X}{\in}\R^{4n \times n_\mathrm{s}}$ consists of the states $\mm{x}_k$ and inputs $\mm{u}_k$ for $n_\mathrm{s}$ samples.}
The states are also passed separately as ground truth $\mm{Y}{\in}\R^{2n \times n_\mathrm{s}}$ to compare each $\mm{x}_{k+1}$ for the next time step with the prediction $\hat{\mm{x}}_{k+1}$.
\highlightred{All input and output variables of the network are scaled between $-1$ and $1$ by using their minimum/maximum values in the dataset\footnote{\highlightred{For the sake of compactness, we do not introduce new symbols for each variable. Min-max scaling is straightforward and must be taken into account during implementation.}}.}
The time series are first divided into partially overlapping sequences $\mm{X}_\ind{seq}$ and $\mm{Y}_\ind{seq}$, which are stored in $\tilde{\mm{X}}, \tilde{\mm{Y}}$ and then split into a training dataset $\mm{D}_\ind{t}$ and a validation dataset $\mm{D}_\ind{v}$. 
Batches $\mm{b}$ of size $n_\ind{b}$ are then formed containing the shuffled sequences of the two datasets.
All individual batches $\mm{b}$ are stored in $\mm{B}$.
For each sequence of a batch, the hidden states of the GRUs are initialized for a window of $n_\ind{w}$ time steps by feeding the measured states~$\boldsymbol{x}_{k}$ and inputs $\boldsymbol{u}_{k}$ into the model and passing on the hidden states.
After this \textit{warm-up phase} (lines \ref{warmupstart}--\ref{warmupend}), the network only sees the inputs $\boldsymbol{u}_{k}$ and its self-predicted states $\hat{\mm{x}}_{k}$, in order to predict $n_\ind{p}$ time steps into the future (lines \ref{predstart}--\ref{predend}).
The hidden states are still passed forward at each time step.
This is intended to \textit{imitate the use in the context of MPC}, where measured values are available at the beginning of the prediction horizon, followed by self-loop prediction into the future.
Only the results of the recursive self-prediction are used to calculate the loss, which leads to a truncated backpropagation through time.
The warm-up phase of the hidden states is necessary because it simulates the use in the MPC with initialized hidden states.
Thus, the network can utilize measured values and then give meaningful self-loop predictions \textit{even for a few time steps in the prediction horizon of the MPC}.
This is contrary to the conventional batch-wise training of RNNs, where all the past state information is stored in the \textit{non-measurable hidden states}.
Even when measured states $\mm{x}_k$ are fed into these RNNs, they are not utilized, which is shown in \ref{sub:rnn_performance}.

The training is carried out with PyTorch using the Adam optimizer.
The loss $\mc{L_\ind{t}}$ that is minimized during training is the mean-square error~(MSE) between the measured and predicted states over an entire batch~$\mm{b}$.
After each epoch, $\eta$ is adjusted with a scheduler (line \ref{LRsched}). This decreases the learning rate if there is no improvement in the total validation loss~$\mc{L}_\ind{v,epoch}$ for~$n_\eta$ epochs, and thus, a plateau exists.
\subsection{Hyperparameter Optimization}\label{sub:hyperparam}
Several hyperparameters have a major influence on the network performance.
\highlightredsec{However, the hyperparameters of LSTM networks and GRUs have never been optimized in the context of soft-robot modeling}.
We use the asynchronous successive halving algorithm~(ASHA)~\cite{Li.2020} to systematically optimize the hyperparameters.
ASHA takes random samples within the specified limits of the hyperparameters and starts multiple training runs based on the available computing resources.
By monitoring the validation loss of each configuration during training, trials with poor performance can be stopped early.
This increases the number of configurations to be tested by several orders of magnitude for fixed computing resources.
\highlightred{Based on the MSE on the validation dataset, the best hyperparameter configuration is selected. To evaluate the generalizability, this model is then tested on an \textit{independent test dataset} according to best practices. More information regarding the different datasets can be found in~\ref{sub:rnn_performance}.}

\begin{figure}[t] 
	\vspace{2.5mm}
	\centering
	\includegraphics[width=.94\linewidth]{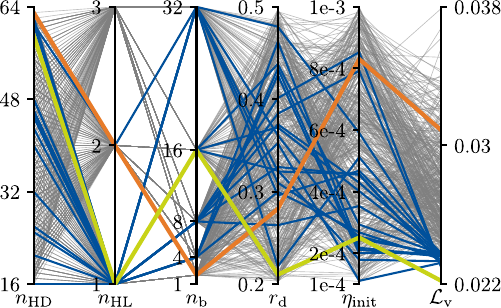}
	\caption{\highlightred{HPO results for the GRUs: Each line represents a trial (combination of $n_\mathrm{HD}$, $n_\mathrm{HL}$, $n_\mathrm{b}$, $r_\mathrm{d}$ and $\eta_\mathrm{init}$). Poorly performing trials are shown in gray, the best twenty in blue, and the best one in green. A baseline configuration is highlighted in orange. The validation loss $\mc{L}_\ind{v}$ is considerably reduced by systematically determining the optimum hyperparameters.}}
	\label{fig:plot_hyperparams}
	\vspace{-2mm}
\end{figure}

The HPO was carried out on a computing cluster \highlightred{(Intel Xeon Gold CPU)}.
For our application, the hidden dimension~$n_\mathrm{HD}$, the number of hidden layers~$n_\mathrm{HL}$, the batch size~$n_\mathrm{b}$, the dropout rate~$r_\mathrm{d}$, and the initial learning rate~$\eta_\mathrm{init}$ were used as hyperparameters.
A grace period of $100$ was used with a maximum of $n_\ind{e}{=}300$ epochs. We set $n_\eta{=}10$, $n_\ind{w}{=}100$ and $n_\ind{p}{=}20$.
\highlightred{Note that the HPO was conducted several times to iteratively determine suitable ranges of all hyperparameters. This ensures that we do not obtain suboptimal results at the border of the parameter space.}

The individual trials of the HPO of the GRUs are shown in Fig.~\ref{fig:plot_hyperparams}, which helps to understand the influence of the various parameters.
\highlightred{For comparison, a baseline configuration with \textit{reasonable} hyperparameters is also shown. 
Using only one configuration is used in most related works such as~\cite{Luong.2021, Hyatt.2019, Li.2022} instead of a comprehensive HPO.}
\highlightred{The best configuration achieved an MSE of $0.022$, which is an improvement of $\approx21\%$ compared to the baseline MSE ($0.028$).}
\highlightred{Qualitatively similar results were achieved in the optimization of the LSTM network, which is not shown due to limited space.}

\highlightredsec{There are two practical remarks to be made: 
First, it is important that the network complexity is not too large to enable real-time control.
The MPC solver would take too long by using an accurate but very complex RNN, which decreases the maximum possible control frequency.
Therefore, the tradeoff between accuracy and evaluation speed of the network must be considered.
Second, the interplay between the network's prediction frequency and the prediction horizon $T$ is crucial, while the latter substantially influences the solving times.
We chose $T{=}4$ to allow real-time control and tuned the prediction frequency accordingly.
A higher prediction frequency shortens the time that the MPC predicts into the future.
This can lead to difficulties for too short prediction horizons due to aggressive MPC actions or convergence issues of the solver.
Also, a too long prediction horizon (very coarse sampling with too small prediction frequency) hinders precise control of the desired trajectory. 
After a few iterations, the prediction frequency of the network was tuned to $\SI{5}{\Hz}$.
It must be adjusted for systems with slower or faster dynamics.
Our choice enables real-time control with a control frequency of $\SI{5}{\Hz}$ for our system, which could be further increased with better computing hardware.
}
\subsection{Nonlinear Model Predictive Control}\label{sub:lb_mpc}
When controlling systems with several DoF, decentralized controllers are often implemented for each DoF.
MPC, in contrast, enables centralized control of several DoF simultaneously.
It uses the discrete model of the system as a dynamic constraint to predict the behavior for a prediction horizon of $T$ time steps.
Combined with state and input constraints, the MPC solver searches for an input trajectory that minimizes the defined cost function over the prediction horizon.
Only the first time step $\mm{u}_1$ of this optimized input trajectory is applied, and the optimization problem is solved again in each time step.
For the ASR, it is formulated as
\begin{multline}\label{eq:optimization_problem}
	\text{minimize} \enspace  \sum_{k=1}^{T-1} \bigl( \lVert \mm{x}_{\text{des},k}-\hat{\mm{x}}_{k} \rVert_{\mm{Q}_\ind{s}}^{2} + 
	\lVert \Delta \hat{\mm{x}}_{k} \rVert_{\mm{Q}_\ind{d}}^{2} + \\
	\lVert \Delta \mm{u}_{k} \rVert_{\mm{R}_\ind{d}}^{2} + 
	 \lVert \Delta \mm{u}_{\text{stiff},k} \rVert_{\mm{R}_\ind{m}}^{2} \bigr)+ 
	\lVert \mm{x}_{\text{des},T}-\hat{\mm{x}}_{T} \rVert_{\mm{Q}_{\ind{t}}}^{2}
\end{multline}
\highlightred{subject to $[\hat{\mm{x}}_{k+1},\mm{h}_{k}] {=} \mm{f}(\hat{\mm{x}}_{k},\mm{u}_{k}, \mm{h}_{k-1})$, $\vert \hat{\mm{x}}_{k} \vert {\leq} \mm{x}_{\text{max}}$, $\mm{u}_{\text{min}} {\leq}  \mm{u}_{k} {\leq} \mm{u}_{\text{max}}$\highlightred{, and $\hat{\mm{x}}_0$ obtained from measurements for each MPC cycle.} Since the RNN $\mm{f}$ uses scaled inputs/outputs, the entire MPC problem is formulated with scaled (unitless) quantities.}
\begin{figure}[t] 
	\vspace{2.5mm}
	\centering
	\def\svgwidth{\linewidth}
	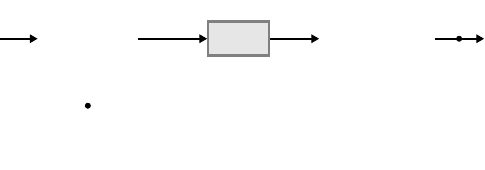
	\caption{\highlightredsec{Block diagram of the learning-based NMPC: An RNN is used as a dynamic constraint to calculate the optimized inputs $\mm{u}$ given the desired state sequence $\mm{x}_\ind{des}$ for the whole prediction horizon. The green network calculates the hidden states\protect\footref{hidden_state_footnote} $\mm{h}$ in each time step, which are passed to the NMPC after a unit delay of $z^{-1}$ together with the current states $\mm{x}$. To prevent confusion, the index for the control time step is omitted, which is not equal to the time step $k$ within the prediction horizon (\ref{eq:optimization_problem}).}} 
	\label{fig:mpc}
	\vspace{-2mm}
\end{figure}
\begin{figure*}[th] 
	\vspace{2.5mm}
	\centering
	\includegraphics[width=.93\linewidth]{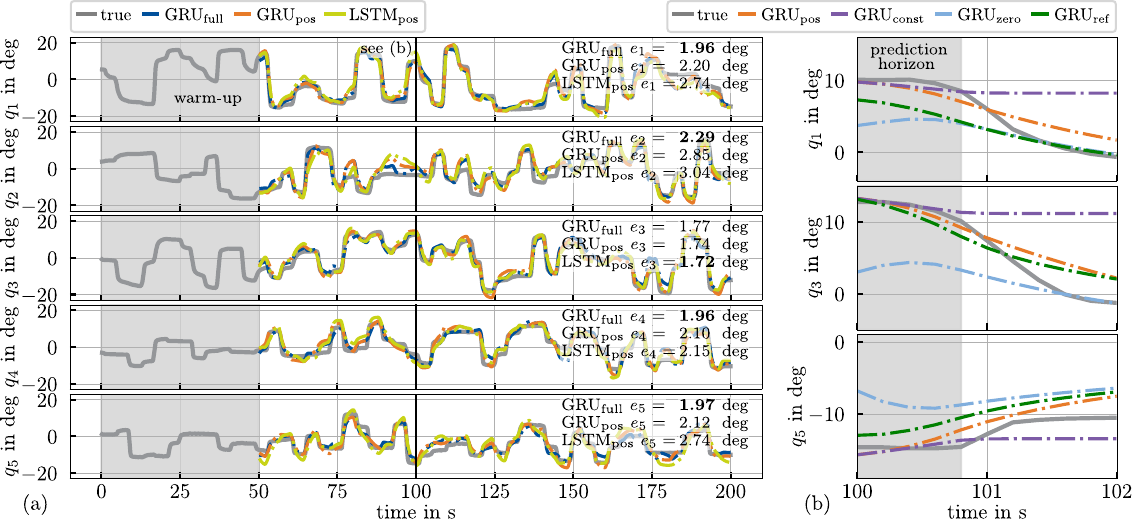}
	\caption{(a) Prediction on test data with root-mean-square error (RMSE) $e_i$.
		RNNs receive measurements to initialize the hidden states (gray area).
		They then predict the further course solely with their outputs and given inputs.
		In contrast to GRU$_{\mathrm{full}}$, which includes the position and velocity as a state, GRU$_\mathrm{pos}$ and LSTM$_\mathrm{pos}$ only uses the position. 
		(b) Performance within short (\SI{0.8}{\second}) prediction horizon. Networks receive measured states at $t{=}\SI{100}{\second}$ and must predict the future course recursively, which simulates the use within MPC. Hidden states of GRU$_\mathrm{zero}$ are naively initialized with zeros. GRU$_\mathrm{pos}$ and GRU$_\mathrm{const}$ receive initialized hidden states, which are available due to the past predictions. Hidden states are kept constant with GRU$_\mathrm{const}$, which still results in high accuracy within horizon. GRU$_\mathrm{ref}$ represents conventionally trained network, and results in larger deviations despite initialized hidden states.}
	\label{fig:plot_prediction}
	\vspace{-2mm}
\end{figure*}
The diagonal weighting matrices $\mm{Q}_\text{s}$, $\mm{Q}_\text{d}$, $\mm{Q}_\text{t}$,  $\mm{R}_\text{d}$ and $\mm{R}_\text{m}$ consist of constant diagonal entries $Q_\text{s}$, $Q_\text{d}$, $Q_\text{t}$, $R_\text{d}$ and $R_\text{m}$.
The input limits are $\mm{u}_{\text{min}}$ and $\mm{u}_{\text{max}}$, and the state limit is $\mm{x}_\text{max}$ for symmetric boundaries.
\highlightred{These can be obtained by min-max scaling of the system-specific limits (pressure range: $0{-}\SI{0.7}{\bar}$, maximum joint angle: $\SI{20}{\degree}$).}
The desired state for the next $T$ time steps is defined as $\mm{x}_\text{des}{=}[\mm{x}_{\text{des},1}\transpose,\dots,\mm{x}_{\text{des},T}\transpose]\transpose$.
With $\mm{Q}_\ind{s}$, the stage cost penalizes deviations from the desired state, whereby a separate terminal cost $\mm{Q}_{\text{t}}$ is set for the last time step.
To penalize the state change, $\mm{Q}_\ind{d}$ is used with $\Delta \hat{\mm{x}}_{k}{=}\hat{\mm{x}}_{k}{-}\hat{\mm{x}}_{k-1}$.
The same is done for the input costs with $\Delta \mm{u}_{k}{=}\mm{u}_{k}{-}\mm{u}_{k-1}$ to place the costs on the pressure change and not on $\mm{u}$ itself.
Thus, $\mm{R}_\text{d}$ can be used to generate smoother pressure curves and avoid oscillations.
In addition,~$\mm{R}_\text{m}$ can be used to keep the mean pressure in both bellows of an actuator at a predefined value $u_\text{mean}$.
For this, we use $\Delta \mm{u}_{\text{stiff},k}{=}\mm{u}_{k1} {+} \mm{u}_{k2} {-} 2[u_{\text{mean}},\dots,u_{\text{mean}}]\transpose$ to adjust the stiffness of the system with $\mm{u}_{k\diamond}{=}[p_{1\diamond,\ind{des}},\dots,p_{n\diamond,\ind{des}}]\transpose$.
\highlightred{This facilitates convergence due to an infinite number of solutions for a desired joint angle given two input pressures.
We choose a mean pressure of $\SI{0.35}{\bar}$, which must be min-max scaled to determine $u_\text{mean}$.}
The hidden states are kept constant within the dynamic constraint to reduce the computational costs for solving the optimization problem.
This simplification is justified in~\ref{sub:rnn_performance}.

A block diagram of the implemented control scheme is given in Fig.~\ref{fig:mpc}.
The MPC solver receives the measured states, desired joint states for $T$ time steps, and the hidden states in each time step and calculates the desired inputs.
The NMPC problem was implemented using CasADi~\cite{Andersson.2019} with the Interior Point Optimizer.

%
\section{Experiments}\label{sec:experiments}
The proposed learning-based MPC is validated using the 3D-printed ASR with cast silicone bellows.
For this purpose, the \highlightred{test bench implementation} is presented (\ref{sub:test_bench}).
The accuracy of the learned RNNs is compared (\ref{sub:rnn_performance}), and control experiments are carried out (\ref{sub:control_results}), which show that the approach enables position control without model knowledge.
\subsection{Test Bench}\label{sub:test_bench}
\highlightredsec{The test bench used to control the soft robot was presented in~\cite{Habich.2024}, and is, therefore, only briefly described below.}

\subsubsection{Architecture}
\highlightred{The pneumatic system consists of a central supply unit and proportional piezo valves with integrated pressure control for each bellows.
The compressed air is generated centrally with industrial compressors and constantly supplied with negligible pressure fluctuations.}
Test bench communication is done via the EtherCAT protocol, which enables \highlightred{several} values (current pressures~$\boldsymbol{p}$, joint angles~$\boldsymbol{q}$ and set pressures~$\boldsymbol{p}_{\text{des}}$) to be read in or set with a cycle time of up to $\SI{1}{\milli\second}$.
\highlightred{The control design is performed on the development computer~(Dev-PC, 3.6~GHz Intel Core i7-12700K CPU with 16~GB RAM) using Matlab/Simulink, and the compiled model is then run on the real-time computer~(RT-PC, 4.7~GHz Intel Core i7-12700K CPU with 16~GB RAM).}

\subsubsection{Implementation of the MPC}
Since CasADi's Matlab API cannot be compiled in Simulink, it was integrated using its \CPP~API. 
For this, the MPC problem is compiled into a shared library and then called in an S-function.
\highlightred{Computation} time is often a bottleneck for NMPC, which makes this implementation even more valuable.
All \highlightred{parameters} of the trained RNN are exported from PyTorch.
The network is manually recreated in CasADi using matrix multiplications. 
\subsection{RNN Performance}\label{sub:rnn_performance}
\begin{figure}[t] 
	\vspace{2.5mm}
	\centering
	\includegraphics[width=\linewidth]{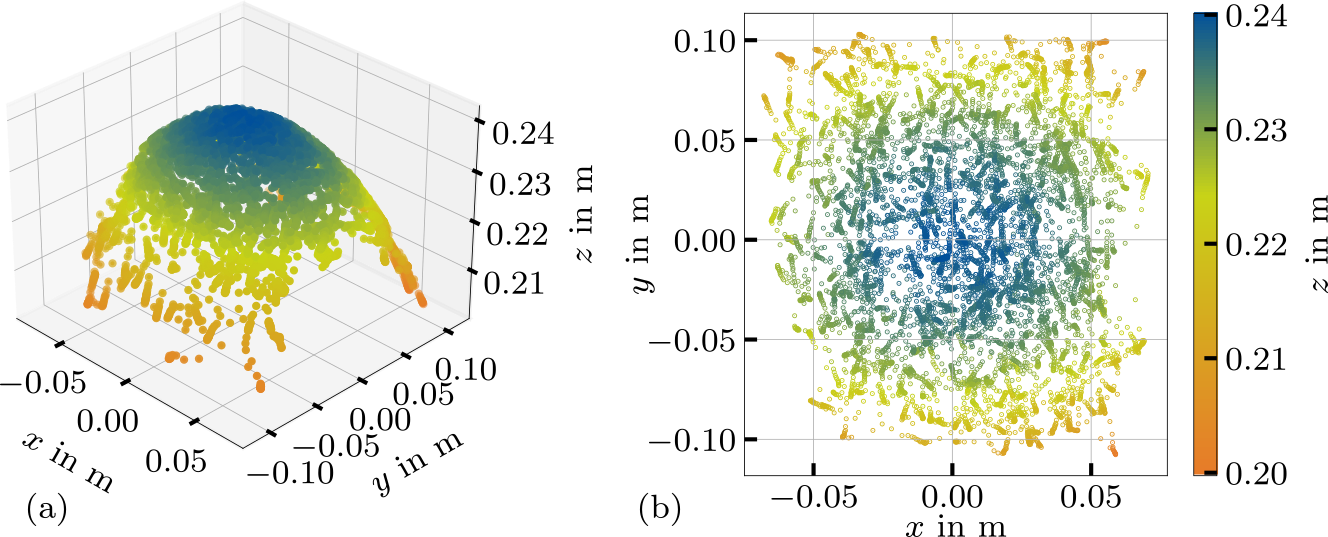}
	\caption{\highlightredsec{End-effector positions in training data: (a) 3D and (b) top view}}
	\label{fig:dataset}
	\vspace{-2mm}
\end{figure}

For the training of the neural networks, two datasets of $\SI{30}{\minute}$ each were recorded using the ASR.
The data was logged at a frequency of $\SI{1}{\kilo\Hz}$ and then downsampled to $\SI{5}{\Hz}$ for the neural networks, resulting in $n_\mathrm{s}{=}9000$ samples.
For the first dataset, random pressure combinations were applied to the actuators as a step, each of which was held for $\SI{4}{\s}$.
This type of system excitation allows more system modes to be stimulated, thus producing a broader range of system responses compared to other common types of excitation. 
\highlightredsec{The end-effector positions are evenly distributed in the task space, which is illustrated in Fig.~\ref{fig:dataset}.}
Random pressure combinations were also used for the second dataset.
However, the transition between these was linear, resulting in ramp-shaped input signals.
\highlightredsec{This also leads to evenly distributed end-effector positions similar to Fig.~\ref{fig:dataset}, which are driven with smoother pressure changes.}
The transition lasted $\SI{1}{\s}$ in each case, and the pressure level was then held for another $\SI{3}{\s}$.
\highlightred{In all experiments, the desired pressures are limited to $\SI{0.7}{\bar}$ to prevent damage.
For other systems, steps may cause damage such that other trajectories must be selected.
If a random selection of the input commands is not possible, these could be selected according to a full-factorial test plan, for example.
Also, a simple PI controller in the joint space~(cf.~IV-C) could be used to move roughly on a safe trajectory.
This is not specific to our work, as it is required for all data-driven identification/learning approaches.}

Of the dataset with pressure steps, $70\%$ was used for training and $30\%$ for validation.
The dataset with pressure ramps was used as \highlightred{an independent} test dataset.
Fig.~\ref{fig:plot_prediction}(a) shows the predictions of the RNNs based on GRUs and LSTM units after a warm-up of their hidden states.
This warm-up time is set to $\SI{50}{\s}$ to ensure that the hidden states are fully initialized.
\highlightred{It was determined empirically, includes a sufficient safety margin, and can vary depending on the system.}
Without this, they initially show poor prediction results.
The GRU$_\mathrm{full}$ (states: $\mm{q}, \dot{\mm{q}}$) is able to model the behavior of the ASR over long periods with an average deviation of $\SI{2.0}{\degree}$.
Using only the position as a state, the GRU$_\mathrm{pos}$ can achieve a deviation of $\SI{2.2}{\degree}$ and the LSTM$_\mathrm{pos}$ has a deviation of $\SI{2.5}{\degree}$.
GRUs are used for the learning-based MPC due to their simpler structure, which is advantageous for nonlinear MPC.
Since using the velocity only slightly improves accuracy and also complicates the structure of the MPC, \textit{only the position is used as a state within MPC}.

For GRU$_\ind{pos}$, the role of the hidden states for short prediction horizons is further analyzed.
The chosen control frequency of $\SI{5}\hertz$ and $T{=}4$ results in a prediction horizon with a duration of $\SI{0.8}{\second}$.
Fig.~\ref{fig:plot_prediction}(b) simulates the \textit{accuracy within the prediction horizon during MPC} at a random time $t{=}\SI{100}{\second}$ for different configurations.
Therefore, measured states are available at the beginning, and the states must be predicted recursively given the system's inputs.
GRU$_\mathrm{zero}$ consists of naively initialized hidden states that are set to zero, which leads to large prediction errors.
GRU$_\mathrm{pos}$ and GRU$_\mathrm{const}$ both receive the hidden states, which are recursively initialized during the past $\SI{100}{\second}$.
The latter keeps the hidden states constant during the self-loop prediction.
GRU$_\mathrm{const}$ shows that maintaining the initialized hidden states constant over a few time steps leads to only a slight reduction in accuracy.
This enables a twice as fast NMPC by maintaining constant hidden states throughout the prediction horizon, which are reinitialized with measured data after each MPC cycle.

As a reference, GRU$_\ind{ref}$ represents a conventional trained GRU without our approach. 
The same hyperparameters and dataset are used during batch training. 
The hidden states are set to zero in the first batch and passed between the batches using gradient detaching.
\highlightred{Over long trajectories, the prediction accuracy of GRU$_\ind{ref}$ is comparable to GRU$_\mathrm{pos}$.
However, for MPC, accuracy within short prediction horizons is crucial.}
It can be seen that the accuracy of GRU$_\ind{ref}$ is low within this short horizon compared to GRU$_\mathrm{pos}$ and GRU$_\mathrm{const}$.
There are large deviations, particularly at the beginning of the horizon, which indicates that the measured states are not being utilized.
\highlightredsec{During control, this deviation prevents the solver from converging.
It was therefore not possible to set up a control system with the conventional GRU$_\ind{ref}$.}

\subsection{Control Results}\label{sub:control_results}
A test trajectory was created to evaluate the accuracy of the learning-based NMPC. 
It consists of ramps with a random slope, between which the position is held briefly. 
All five actuators are moved simultaneously.
\highlightred{All weighting matrices are tuned manually for good tracking performance.}
The entries of the matrices $\mm{Q}_\text{s}$ and $\mm{Q}_\text{t}$, which sanction the deviation from the desired position, are set to $Q_\text{s}{=}5$ and $Q_\text{t}{=}10$.
$\mm{Q}_\text{d}$ can be used to influence the speed of the robot without using $\dot{\mm{q}}$ as a state.
As this was not relevant to the experiment, it was set to $Q_\text{d}{=}0$.
In order to obtain smoother pressure curves, the entries of the matrix $\mm{R}_\text{d}$ were set to $R_\text{d} {=} 4$.
To achieve a uniform mean stiffness of the ASR, the entries of the matrix $\mm{R}_\text{m}$ were set to $R_\text{m} {=} 5$.

\begin{figure}[t] 
	\vspace{2.5mm}
	\centering
	\includegraphics[width=\linewidth]{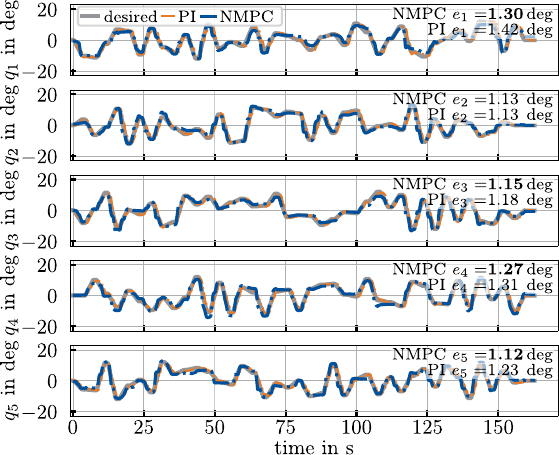}
	\caption{Trajectory tracking and RMSE~$e_i$ during position control using the learning-based NMPC \highlightred{and comparison with PI control}} 
	\label{fig:plot_control_all}
	\vspace{-2mm}
\end{figure}

The results of the position control are shown in Fig.~\ref{fig:plot_control_all}.
Our approach demonstrates a $5\%$ improvement in accuracy compared to a PI controller, which is based on~\cite{Habich.2024}. It must be mentioned that the PI controller runs at $\SI{1}{\kilo\hertz}$, and the control frequency of the MPC is currently limited to $\SI{5}{\hertz}$. \highlightredsec{Unlike the PI controller, the MPC performance could be further increased by using better computing hardware (CPU and GPU) to realize higher control frequencies. Also, note that this improvement strongly depends on the system dynamics and can be higher for different systems due to the feedforward/feedback character of MPC.
Another advantage is constraint satisfaction, e.g., to influence the maximum speed of the robot.}
Overall, the learning-based NMPC is able to reliably control the ASR with an average tracking error of around $\SI{1.2}{\degree}$.
It reacts fast to changes in the target position, as it predicts a few steps into the future.
Occasionally, deviations occur due to model uncertainties, which are mainly caused by static friction effects that are difficult to predict.
%
\section{Conclusions}\label{sec:conclusions}
We present a universal approach for learning-based nonlinear model predictive control of soft robots based on recurrent neural networks.
To use an RNN within NMPC, the focus must be placed on the correct handling of the hidden states. 
For this purpose, \highlightred{we propose a training approach that enables high accuracy in short prediction horizons}.
A discrete-time nonlinear model of the multi-DoF ASR was trained with RNNs while optimizing the hyperparameters.
Comparisons between LSTM and GRU networks revealed that GRUs achieve better accuracy.
Real experiments with the soft robot demonstrate high model accuracy and accurate trajectory tracking.
No expert knowledge or assumptions about the model are required, allowing this method to be applied to any model-based control problem.
\highlightred{To enhance reproducibility, the entire codebase for learning and control is available as open source\footref{foot:sponge}.}

Future work should focus on hybrid modeling approaches that incorporate both physics-driven and learning-based methods to increase the generalizability of the underlying soft-robot model for changed system dynamics.
An online-learning approach could also continuously update the data-based model to ensure adaptability during wear or replacement of components.
\highlightredsec{Instead of manually tuned controller gains, automated tuning could further improve the control performance.}
\highlightred{Comparing the used RNNs with other networks, such as NARX or transformer, could also be investigated in order to determine the most suitable architecture.}
Transformers show good accuracy in many applications and have no hidden states like GRUs, which simplifies the MPC problem and enables higher control frequencies~\cite{Park.2023}.

 \section*{Author Contributions}
 HS and TLH developed the data-driven models with support from CM and SFGE. HS implemented the MPC and carried out the experiments under TLH's guidance. TLH has conceptualized the core idea of the article. All authors contributed to the manuscript. MS supervised the research.
\section*{Acknowledgment}
We thank Max Bartholdt for his support in implementing the MPC on the test bench.
\bibliographystyle{IEEEtran}
\bibliography{literatur.bib}

\end{document}